# Edge-Adaptive Hiera-Seg Model for Building-Integrated Photovoltaic Power Forecasting

Na Yu[a]

[a]First Highway Engineering Co., Ltd. Chengdu 610000, Sichuan , China

**Abstract**

Building integrated photovoltaic power forecasting is challenging because facade orientation, rooftop geometry, nearby buildings, vegetation, cloud movement, module temperature, inverter status and edge device reliability jointly affect power generation. This paper proposes an edge-adaptive HieraSeg model for short term BIPV power forecasting with multi-orientation visual-irradiance fusion. RGB facade images, optional thermal cues, irradiance measurements, module temperature, facade orientation, rooftop masks, shading labels, sky condition, historical power and edge device state are aligned into synchronous forecasting windows. HieraSeg extracts hierarchical regions from facade, rooftop, sky, shadow and background images, while the edge-adaptive module changes feature weights under missing frames, latency shifts and local shading changes. The segmentation branch identifies occluded module areas and separates facade, rooftop, sky and background evidence before the forecasting head estimates future BIPV output. Experiments are designed on a simulated multibuilding BIPV dataset with unseen building and unseen orientation tests. The model reports predicted power, confidence interval, dominant evidence, occlusion contribution, orientation contribution and energy-management review priority. This output is intended to show whether the forecast is driven by irradiance change, visible shading, orientation response, stale edge information or recent power history. The framework supports local forecasting at edge controllers with limited communication delay and computing resources, linking visual occlusion, irradiance dynamics and power history to traceable forecasting results for building energy management, microgrid scheduling and BIPV operation planning.

## 1. Introduction

Building-integrated photovoltaic systems have become an important form of distributed solar generation in dense urban buildings [1], because facades and rooftops can produce electricity without requiring additional land. Short-term PV forecasting studies [2] show that irradiance fluctuation, temperature variation and cloud movement strongly affect output stability. BIPV research [3] further indicates that facade modules have more complex generation patterns than open-field PV arrays because their orientation, tilt and surrounding obstruction differ across the same building. Vision-assisted solar forecasting studies [4] also suggest that sky images, module images and shading evidence can improve prediction when irradiance sensors alone cannot represent local occlusion. These findings motivate a model that can combine multi-orientation visual evidence with physical monitoring variables. For BIPV systems, the same global irradiance level can correspond to different power outputs if one facade is shaded by a nearby building and another receives direct sunlight. Therefore, the prediction window should include visual occlusion, facade orientation and edge-device state rather than relying only on historical power.

Most traditional BIPV forecasting methods rely on numerical weather prediction, historical power and module temperature. These sources are useful but cannot fully explain urban shading caused by balconies, adjacent buildings, trees and facade protrusions. Deep learning models improve nonlinear time-series forecasting [5], and segmentation-based visual models can extract shaded and illuminated module regions from images [6]. The revised introduction therefore distinguishes BIPV forecasting from open-field PV forecasting: facade orientation, rooftop shape and urban occlusion change the local power response, and an edge controller may not receive every image frame in time.

Hiera-Seg is proposed as a hierarchical segmentation and forecasting framework. The segmentation branch separates facade modules, rooftop modules, sky regions, shaded areas and irrelevant backgrounds, while the forecasting branch combines these masks with irradiance, temperature and power history. Edge-adaptive modules adjust fusion according to missing frames, inference latency and sensor reliability. The revised text also clarifies that the output is a forecasting package containing predicted power, confidence and dominant evidence, so the building operator can decide whether to use the forecast, delay control or request manual confirmation.

The rest of this paper is organised as follows. Section 2 reviews multi-orientation BIPV generation, urban shading effects and vision-assisted solar forecasting. Section 3 introduces visual-irradiance window alignment and the orientation-aware Hiera-Seg forecasting model. Section 4 presents the simulated dataset, multi-orientation forecasting results, segmentation ablation, robustness analysis and review-action interpretation. Section 5 summarizes the findings and discusses future validation with real BIPV meter data, camera-quality control and edge-device recalibration.

## 2. Related Work

### 2.1. Multi-orientation BIPV Generation and Urban Shading Effects

BIPV generation differs from traditional PV generation in that the modules are integrated into the building envelope. Rooftop modules generally receive stronger direct irradiance, and facade modules may be oriented vertically or at an angle, and are subject to reflection and partial shading. Research on urban PV deployment [7] has shown that building shape and nearby buildings are prone to significant spatial differences in power generation. Therefore, a single irradiance sensor at the roof may not be suitable for facade-level forecasting in all cases. The South facade, the East facade and the West facade may have different peaks in the morning and evening on the same day.

Urban Shading is also a reason. Nearby buildings, trees, railing structures and facade protrusions are moving shadows that change rapidly as the sun moves. Studies on PV shading and mismatch losses [8] have shown that partial occlusion can cause a non-linear reduction in output due to altered electrical characteristics of module strings. Shading in BIPV systems may be localised and repeated, such as a balcony shadow at the same time each day, or transient and irregular, such as cloud movement. Therefore, the forecasting model should consider both the reduction in irradiance and the location and kind of occlusion.

Multi-orientation forecasting also has a different assessment standard. A small absolute error in the rooftop output may be due to a large proportion of rooftop power. Building Energy Management Systems need forecasts for each orientation because storage scheduling, load shifting and facade control actions are often based on local generation. Therefore, the proposed model reports both the total power and the orientation contribution. Therefore, the energy controller can determine whether the cause of the forecast error is a change in rooftop irradiance, a facade shading phenomenon or an unpredictable visual frame. Orientation-aware reporting is also convenient for BIPV output connected to a building microgrid. The controller will determine whether the uncertainty will affect a high-output source requiring a change in battery scheduling or a low-output facade source only subject to local monitoring.

### 2.2. Vision-assisted Segmentation for Solar Power Forecasting

Vision-assisted solar forecasting has been carried out using sky images, module images and remote-sensing images. Image-based models can observe cloud cover, changes in irradiance and local shadow patterns that are not directly accessible at point sensors. Segmentation method [9] is suitable for the forecasting task of separating modules, shadows, skies and backgrounds. Raw images in a BIPV facade contain windows, wall materials, vegetation and neighbouring buildings. Without segmentation, the neural model will learn the background texture instead of power-related occlusion evidence.

Hierarchical division is suitable because solar evidence exists at multiple scales of space. A small shadow in a module cell will reduce the local current, and a large building shadow will reduce the output of an entire facade. Hiera-Seg-style representation can connect local pixel regions, module-level masks and building-level orientation evidence. Therefore, the visual branch is more suitable for BIPV forecasting than a flat image encoder. At the same time, Power forecasting still needs to consider time-series data because the impact of shading changes with fluctuations in power, temperature and irradiance.

Edge deployment has practical deficiencies. Energy controllers may use a local gateway that has a small amount of memory and processing power. If all of the images need to be sent to a cloud server, the communication delay will be too long for timely scheduling. Fan et al. [10] proposed a distributed photovoltaic short-term power forecasting model based on lightweight AI for edge computing in low-voltage distribution networks, showing that local lightweight inference can improve deployment feasibility when communication and computing resources are constrained. This paper takes the above direction and uses edge-adaptive modules to change the fusion weight when visual frames are missing, inference latency increases or sensor reliability decreases.

Based on the above research, a high-accuracy BIPV forecasting model should integrate visual segmentation, orientation information and edge-aware temporal prediction. Existing PV forecasting methods are generally good at time-series modelling but lack local visual occlusion. The current segmentation model can identify regions but fails to convert them into high-priority power forecast and review areas. The proposed Hiera-Seg model connects the above components by using segmentation evidence as input for forecasting and as an explanation of energy-management review. The user of the model is not only interested in an error value but also in this connection. A building operator needs to know if the forecast is uncertain because of visual occlusion, cloud transition, orientation mismatch or edge-device delay. The model should also indicate whether a low-confidence forecast can still be used for regular monitoring or if an automatic dispatch action should be withheld. This review-oriented requirement will shape the output design in the following method section. It also explains why this paper assesses confidence calibration and dominant evidence instead of presenting only the mean forecasting error.

## 3. Methodology

### 3.1. Multi-orientation Visual-irradiance Window Alignment

The proposed framework begins with a multi-orientation visual-irradiance window. Each window contains RGB facade images, optional thermal cues, irradiance measurements, module temperature, facade orientation, rooftop mask, shading label, historical power and edge-device state. Figure 1 shows the multi-orientation BIPV visual-irradiance input framework; it aligns real-scene images, segmentation cues and sensor sequences before the data enter the forecasting model. Late visual frames and missing sensor points are explicitly marked rather than silently removed.

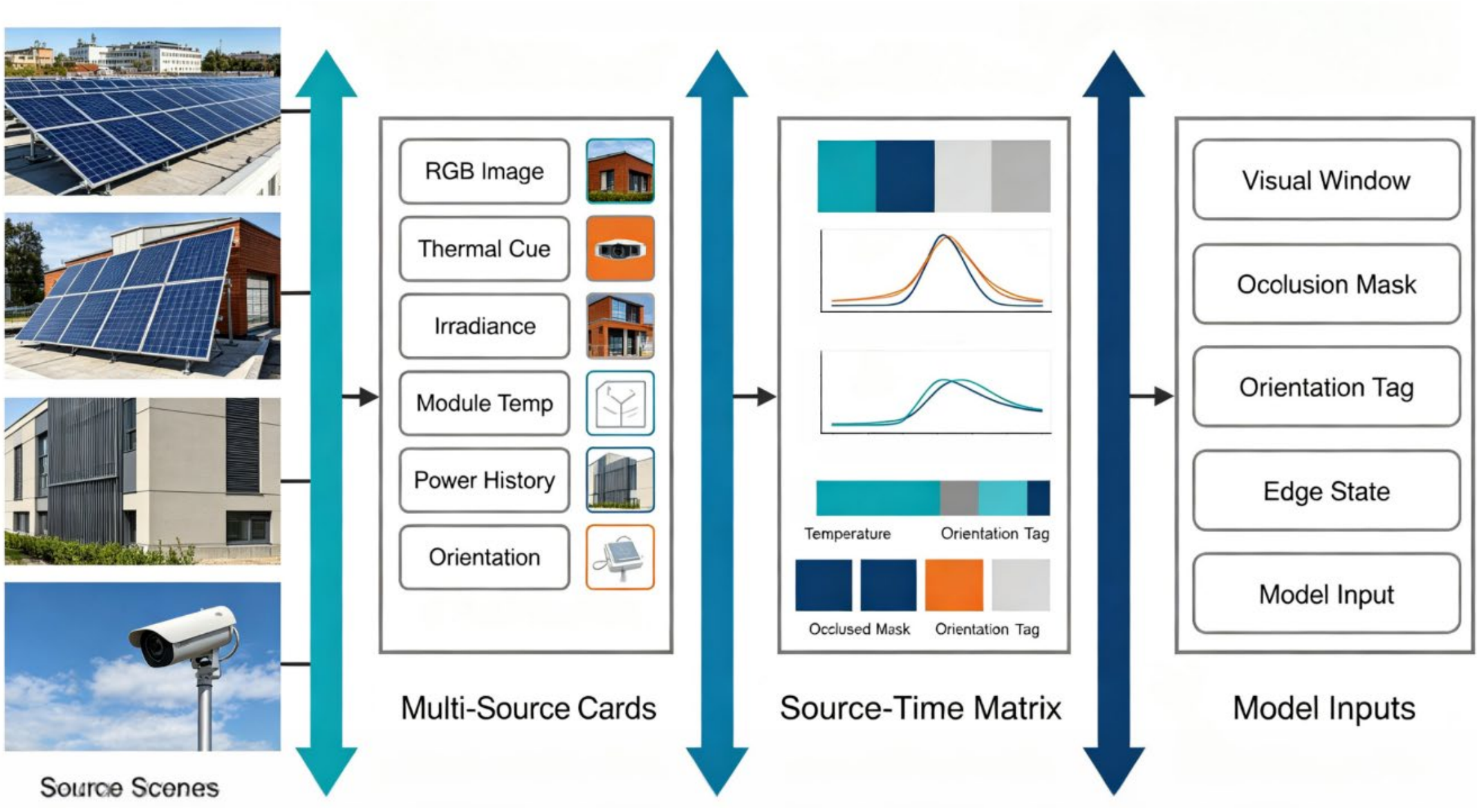

**Figure 1. Multi-orientation BIPV visual-irradiance input framework.**

$$X_t = [I_t, R_t, T_t, O_t, P_{t-k:t}] \quad (1)$$

In this expression, $X_t$ represents the complete input window at time $t$. $I_t$ denotes visual image evidence, $R_t$ denotes irradiance sequence, $T_t$ denotes module temperature, $O_t$ denotes orientation state and $P_{t-k:t}$ denotes recent power history. Keeping these components together allows the model to learn whether a power drop is caused by irradiance change, local occlusion or orientation-specific response. The window is updated on the edge device, so late image frames and missing sensor points are explicitly marked instead of silently removed.

$$M_t = \mathrm{Seg}(I_t) \quad (2)$$

The segmentation map $M_t$ separates module regions, shadow regions, sky, facade background and irrelevant objects. This step is important because a BIPV image contains many visual elements that do not directly affect power generation. The segmentation map helps the model focus on module and shading evidence. It also provides a visible explanation source: if the forecast changes because the shadow mask expands, the operator can review the corresponding image region.

$$S_t = \frac{\mathrm{Area}(\text{shadow})}{\mathrm{Area}(\text{module})} \quad (3)$$

The shading ratio $S_t$ measures the proportion of module area that is affected by visible shadow. It is not used as the only predictor, but it provides an interpretable bridge between visual segmentation and power forecasting. A high shading ratio does not always imply the same power loss because the affected module string, orientation and irradiance level also matter.

$$F_o = \mathrm{concat}(M_t, R_t, T_t, O_t) \quad (4)$$

The orientation-aware feature $F_o$ combines segmentation evidence with irradiance, temperature and orientation. This feature is used to represent why two module groups under the same irradiance may produce different power because their orientation and shadow state differ. It also prevents the forecasting head from treating all modules as if they shared the same sun-facing geometry. The feature is stored at the window level so that it can be updated when the edge device receives a new image or when the irradiance sensor reports a sudden change.

### 3.2. Hiera-Seg Output Forecasting and Orientation-aware Fusion

The forecasting model contains a hierarchical segmentation encoder, an orientation-aware fusion block, an edge-adaptive module and a multi-task forecasting head. Figure 2 presents the orientation-aware Hiera-Seg output forecasting network. It shows how visual masks, sensor sequences and recent power are fused into predicted BIPV output, confidence and dominant evidence. The hierarchical encoder separates module, shadow, facade, rooftop, sky and background evidence so that the forecasting head does not learn irrelevant wall or vegetation textures as power features.

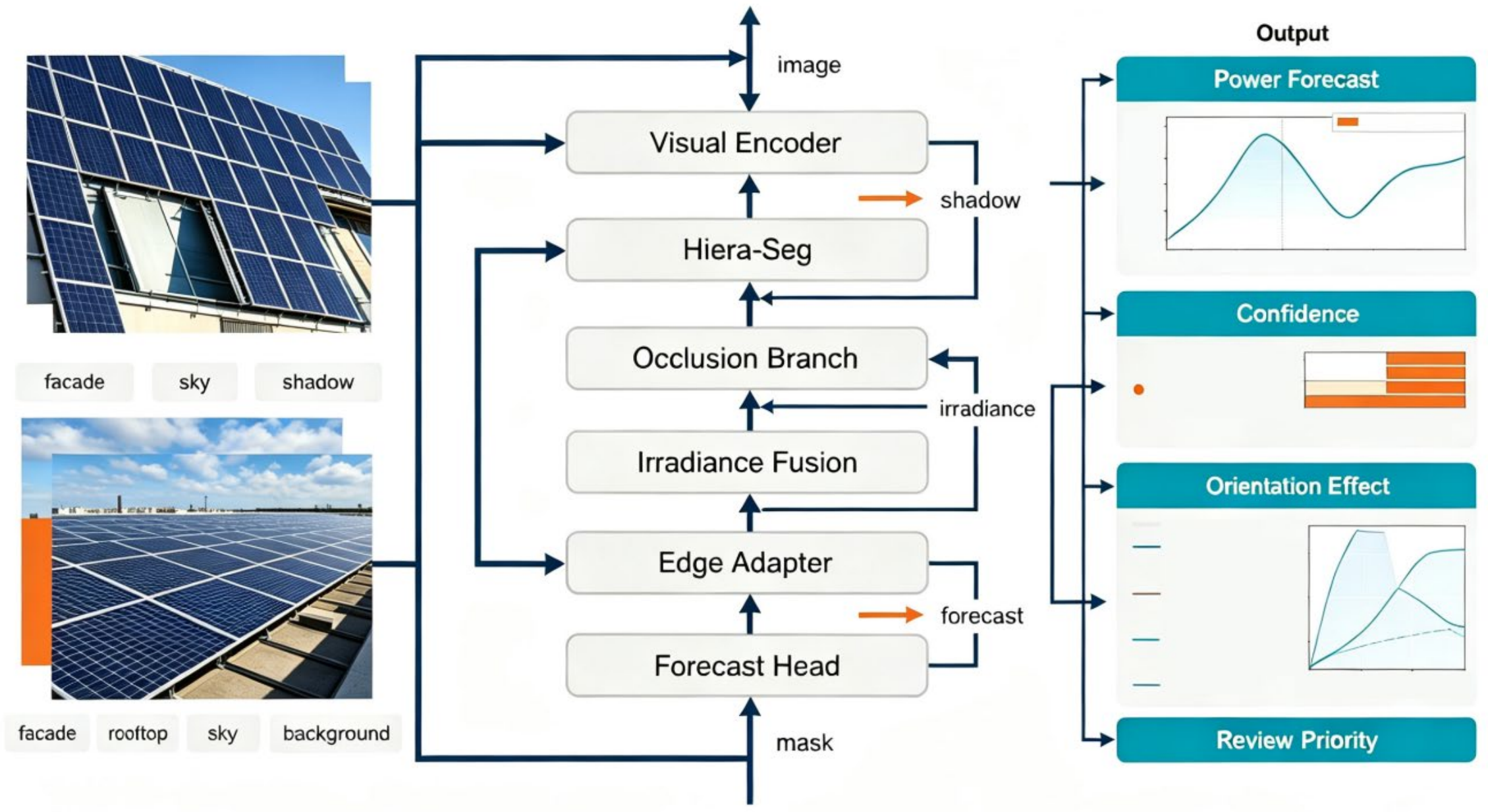

**Figure 2. Orientation-aware Hiera-Seg output forecasting network.**

$$H_1 = \mathrm{Enc}_{\mathrm{local}}(M_t) \tag{5}$$

The local encoder $H_1$ extracts module-level and shadow-level visual features. Local visual evidence is necessary because a small but high-impact shadow can influence a module string even if the whole image appears mostly clear. The local encoder therefore emphasizes boundary regions between illuminated and shaded module areas.

$$H_2 = \mathrm{Enc}_{\mathrm{global}}(H_1, O_t) \tag{6}$$

The global encoder $H_2$ combines local visual evidence with orientation state. This step allows the model to learn facade-level and rooftop-level generation patterns at the same time. It also helps the model separate a real orientation response from a temporary local occlusion pattern.

$$Z_a = \mathrm{Gate}_{\mathrm{edge}}(H_2, E_t) \tag{7}$$

The edge-adaptive gate uses edge-device state $E_t$, including missing frame flag, latency and sensor reliability. When the visual stream is unreliable, the model increases the relative importance of irradiance and power history. When visual evidence is reliable, segmentation features are emphasized. This gate makes the forecasting process more robust than a fixed fusion rule.

$$\hat{P}(t+h) = \mathrm{Head}(Z_a, P_{t-k:t}) \tag{8}$$

The forecasting head predicts future BIPV output at the selected horizon by combining edge-adapted visual evidence with recent power history. The same feature vector generates confidence and dominant-evidence outputs for review. If the image stream is delayed, the model can still forecast through the sensor branch, but confidence is reduced so the downstream controller acts conservatively. This design supports continuous local operation because evidence available at the time of decision is saved with the forecast.

## 4. Results and Discussion

### 4.1. Simulated Dataset and Orientation-shading Distribution

A simulated multi-building BIPV dataset is employed for the proposed model test. The dataset includes South facade, East facade, West facade, rooftop and mixed-orientation arrays in sunny, cloudy, partial-shading and cloud-shade transition conditions. Each record contains RGB visual patches, segmentation masks, irradiance, module temperature, historical power, orientation tags and edge-device state. Because unseen-building generalization is necessary for practical deployment [11], some building facades and rooftops are kept out of training and used only for testing.

Figure 3 shows visual occlusion and irradiance-condition distribution. Figure 3(a) shows visible occlusion area expanding from small facade regions to larger module groups under different building geometries. Figure 3(b) shows irradiance level and separates direct sunlight, diffuse light and cloud-transition windows. Figure 3(c), module temperature, shows that thermal response does not always follow irradiance immediately. With both visual and sensor evidence [12], Figure 3(d) shows clear, cloudy and mixed-sky conditions for mixed-weather evaluation.

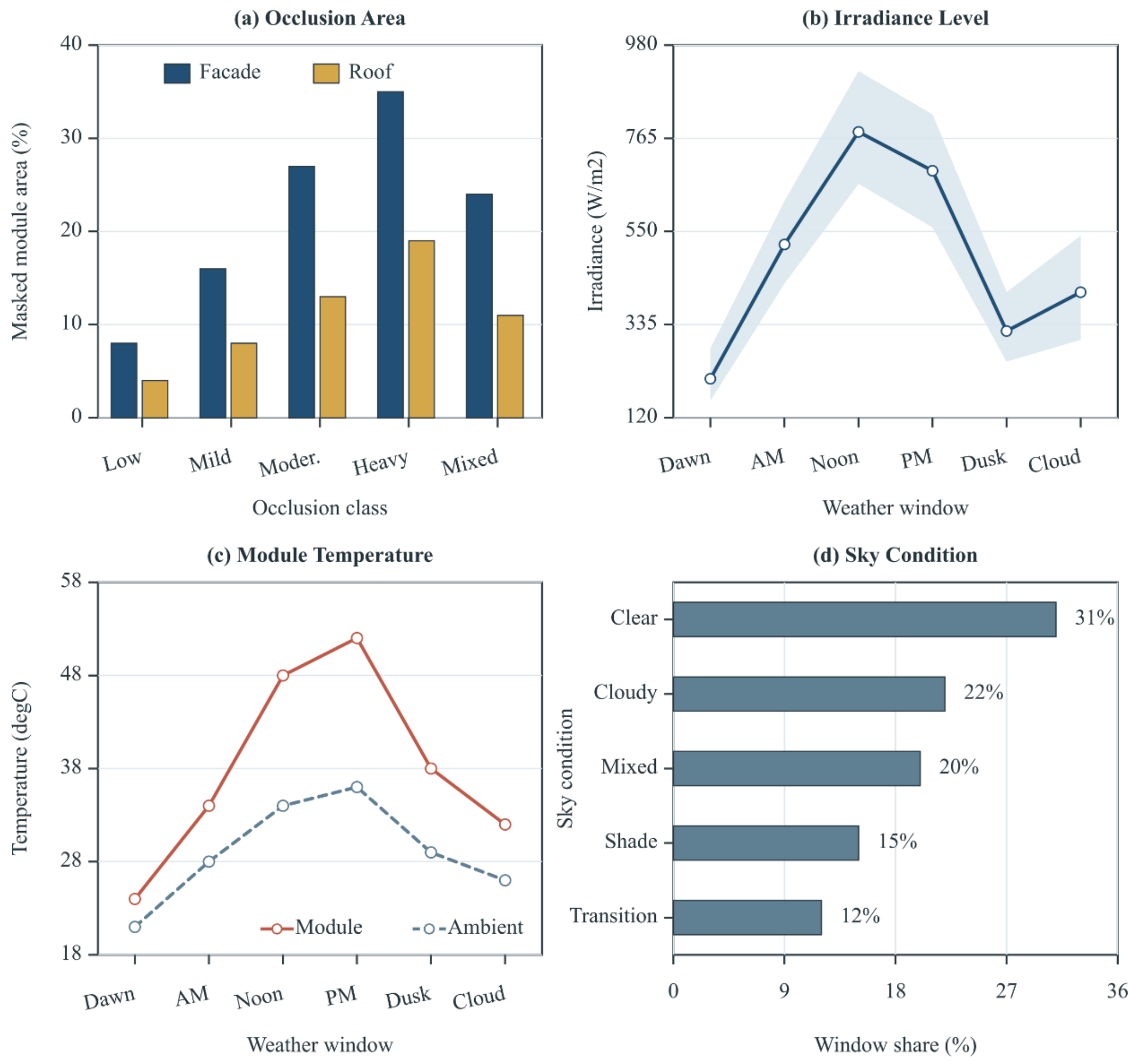

**Figure 3. Visual occlusion and irradiance-condition distribution. (a) Occlusion Area. (b) Irradiance Level. (c) Module Temperature. (d) Sky Condition.**

Figure 4 shows orientation and power-label distribution. Figure 4(a) Orientation group compares south facade, east facade, west facade and rooftop records, showing that generation profiles differ even within the same building. Figure 4(b), Power label, shows the short-term output range and explains why one model must handle both high-output rooftop windows and low-output shaded facade windows. The two subfigures illustrate the transfer problem because orientation and shading jointly alter the relationship between irradiance and output.

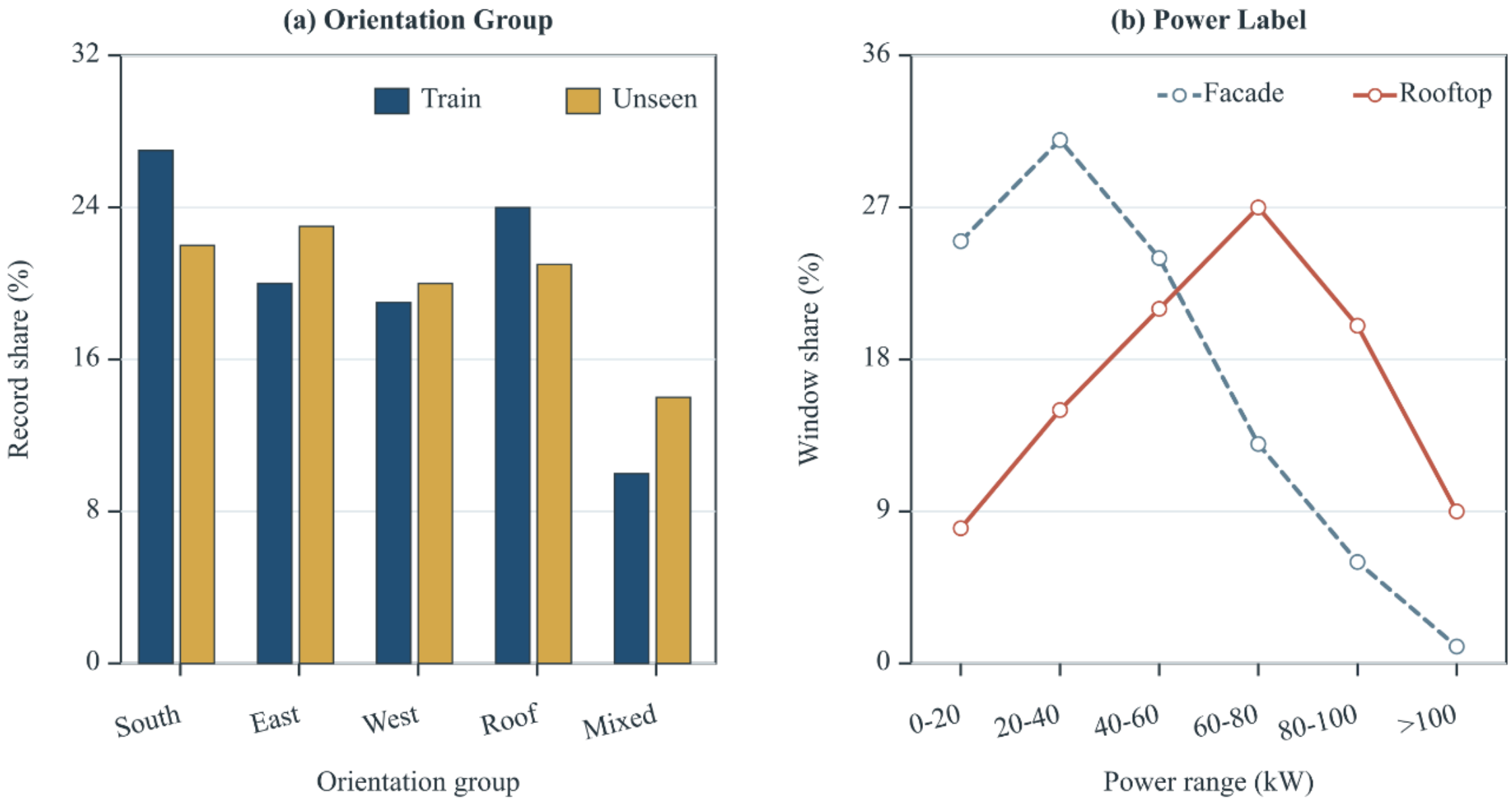

**Figure 4. Orientation and power-label distribution. (a) Orientation Group. (b) Power Label.**

The dataset also has edge-device cases. Some records are missing visual frames, have delayed sensor timestamps or reduced image resolution. Therefore, an edge controller may not always have complete visual information and is used. Therefore, the forecasting model should still have a reasonable forecast even if the segmentation branch is partially uncertain. At the same time, the confidence output should decrease due to missing data and thus be less reliable.

Based on the distribution analysis, BIPV forecasting cannot be reduced to an irradiance-only problem. A rooftop array under uniform light may follow a stable power curve, while a facade with moving shadows can show sudden local variation. Visual occlusion is especially important for facade modules because nearby buildings and facade structures create repetitive orientation-specific shading. The revised discussion therefore separates cases to test whether the model learns local visual evidence or only reproduces average power trends.

Another kind of feature is an overlap of clouds and geometric shadows. A cloudy sky reduces the entire-sky irradiance, and a building shadow only covers a part of the module. The two cases may have the same short-term power reduction but lead to different future behaviour. A cloud will pass, and then there will be irradiance again; however, a geometric shadow will remain until the sun moves. Therefore, for the model to predict the next period accurately, both sky conditions and segmented occlusion evidence are needed.

### 4.2. Multi-orientation Forecasting and Baseline Evaluation

The proposed model is compared with statistical forecasting baselines, recurrent sequence models, Transformer-based forecasters, image-only segmentation models, irradiance-only models and Hiera-Seg variants without edge adaptation. When visual occlusion changes rapidly [13], evaluation reports mean absolute error, root mean square error, orientation-wise error, confidence calibration and unseen-building performance. This design checks whether segmentation evidence is used before the power drop is fully visible in the historical output sequence.

Figure 5 shows multi-orientation forecasting performance comparison. Figure 5(a) is the overall error and reports whether the proposed model improves total forecasting. Figure 5(b) is the facade error, where vertical modules are more exposed to local shadow and reflected light. Figure 5 (c), rooftop error, tests performance on conventional high-output arrays. Figure 5(d) Unseen orientation evaluates whether the model can forecast a facade orientation not fully represented in training. The four subfigures show whether improvement comes from real visual-orientation representation rather than one dominant scenario.

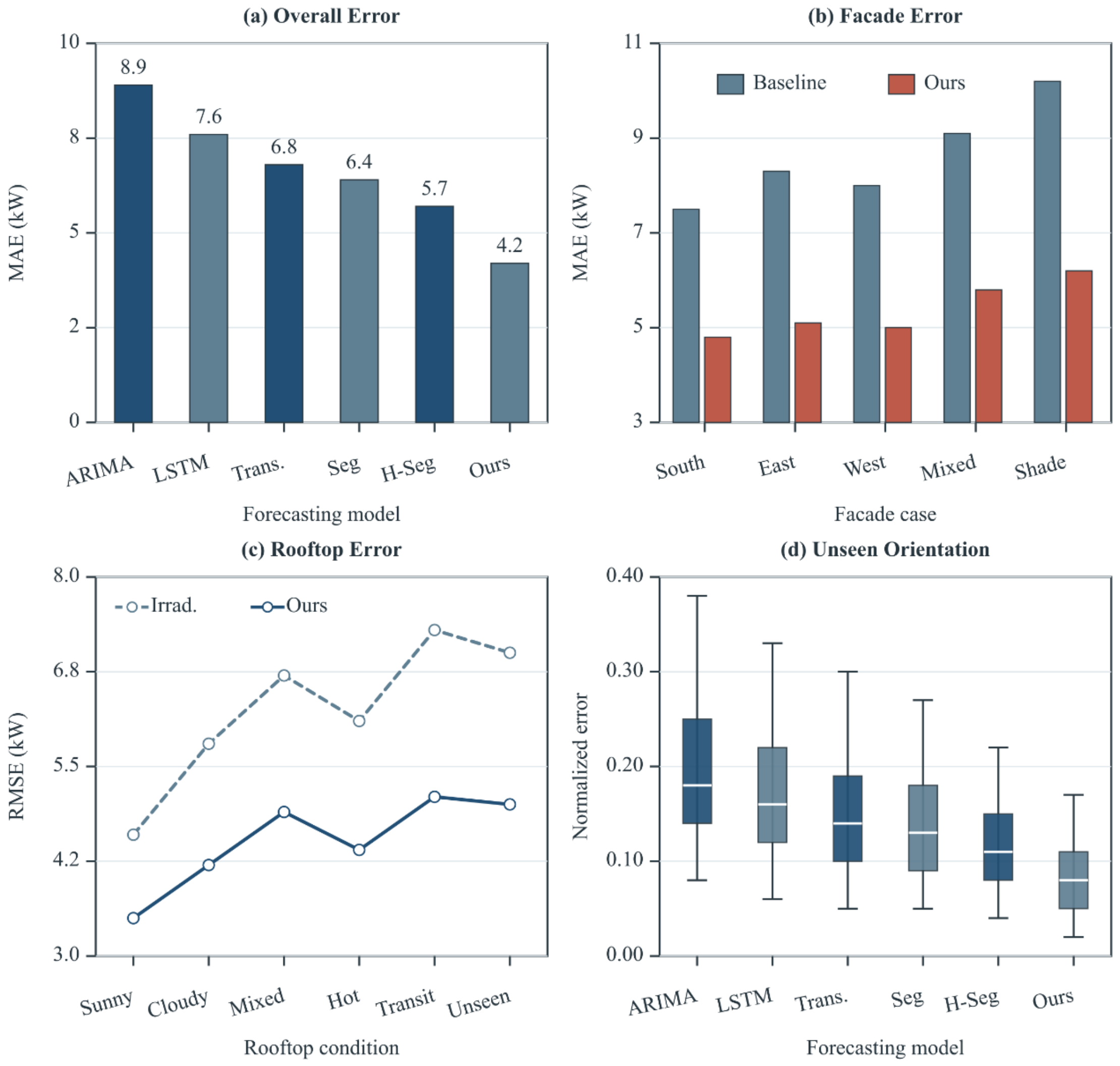

**Figure 5. Multi-orientation forecasting performance comparison. (a) Overall, Error. (b) Facade Error. (c) Rooftop Error. (d) Unseen Orientation.**

Visual segmentation is especially useful in partial-shading environments. Clear-sky rooftop windows can often be forecast from historical power and irradiance, but facade windows require occlusion masks to indicate power reduction before the power sequence fully reacts. Because image-only models may confuse background walls with modules [14], the Hiera-Seg encoder isolates photovoltaic regions and shaded subregions before forecasting. This explains why segmentation is a forecasting component rather than only a visual explanation.

Orientation-wise error analysis shows the specific distribution of the error. South-facing facades have relatively even midday generation, east-facing facades have a strong morning peak, and west-facing facades have a strong afternoon response. Rooftop arrays are less affected by building-plane orientation but more sensitive to cloud cover. When orientation evidence is removed [15], the model tends to underpredict vertical facade output in low-angle sunlight and overpredict shaded facade output due to nearby-building obstruction. Orientation is therefore a structural variable rather than a category label.

The Model also increases short-term stability. Recurrent and Transformer baselines can fit the recent power well, but they sometimes react after the occlusion has already caused a power drop. The visual branch is earlier; therefore, a shadow has appeared before the full-power trend has been formed. Under rapidly changing cloud cover [16], the Hiera-Seg forecast tracks the drop and rise of irradiance more closely than the irradiance-only model. Therefore, the prediction will be inaccurate and may lead to an incorrect charging or discharging time for the battery.

Figure 6 shows visual occlusion sensitivity and segmentation ablation. Figure 6(a) shows how forecast error changes as visible shadow area increases from light obstruction to heavy obstruction. Figure 6(b) segmentation ablation compares full segmentation, module-only segmentation and no-segmentation settings. Segmentation separates relevant module evidence from irrelevant background [17], so the ablation demonstrates whether Hiera-Seg contributes directly to forecasting rather than merely adding an image-processing front end.

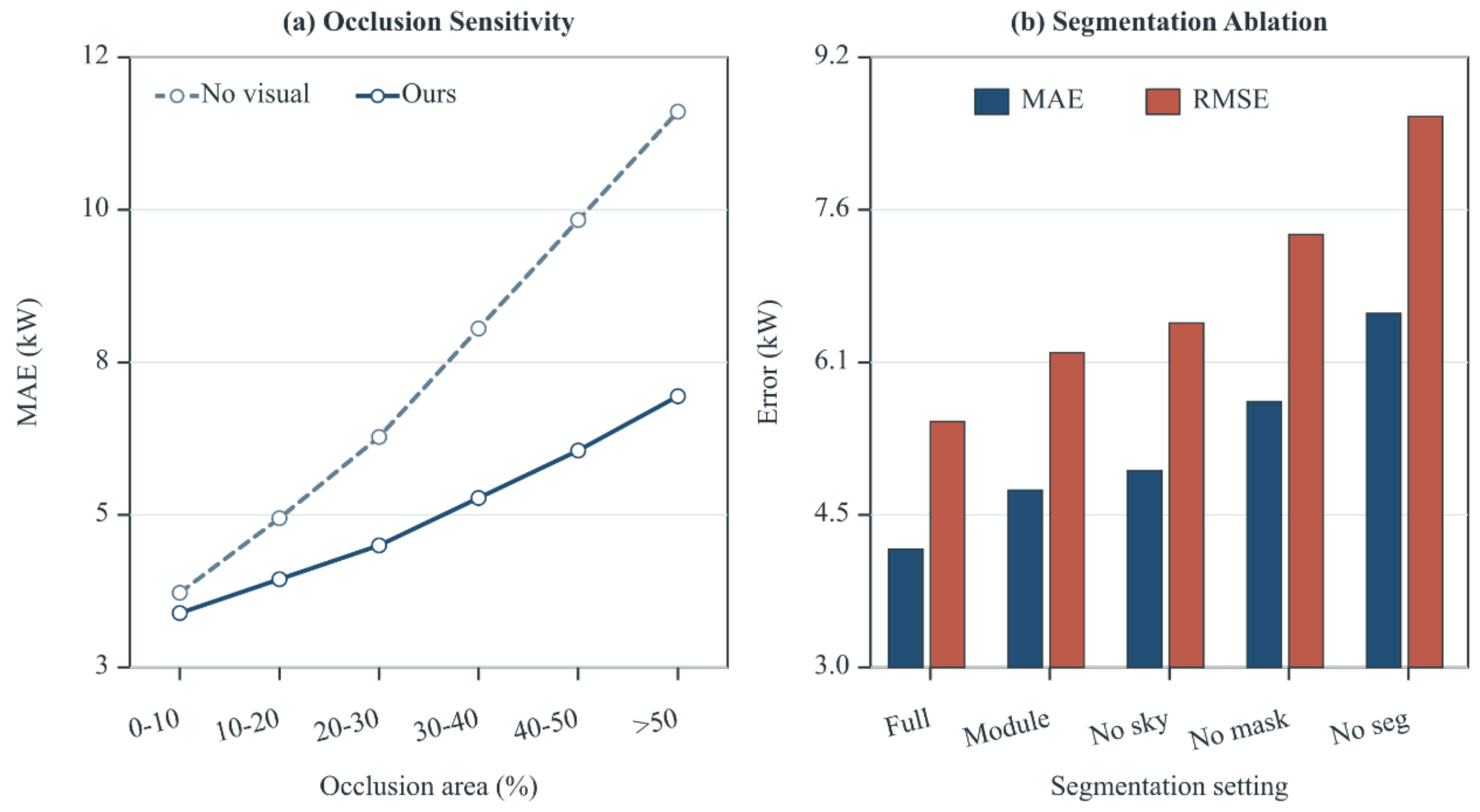

**Figure 6. Visual occlusion sensitivity and segmentation ablation. (a) Occlusion Sensitivity. (b) Segmentation Ablation.**

Based on the ablation results, removing orientation information mainly affects facade forecasts, while removing segmentation mainly affects shaded windows. Edge adaptation has a smaller effect in complete-data cases but a larger effect when image frames are delayed or missing. Under unstable edge inputs [18], orientation provides physical context, segmentation provides local visual evidence and edge adaptation regulates feature-stream reliability. This separation clarifies why the three modules are complementary.

A particular error analysis shows that the difficult cases are concentrated near the transitions. Morning facade activation, sudden cloud movement, partial module shading and reflected-light changes are more difficult than in a stable clear-sky condition. The two are combined to reduce the error of the proposed model. When both the visual frame and irradiance sensor are uncertain [19], the model reduces confidence and assigns the forecast to review instead of outputting a high-confidence result.

The forecast results also have a schedule. A small absolute error in the low-output facade period will have a limited operational impact; however, an error during a peak rooftop period will affect battery charging or load shifting. Therefore, the model output includes confidence and dominant evidence in addition to predicted power. For storage scheduling and controllable-load decisions [20], it helps the building energy-management system decide whether to use the forecast directly or check before acting.

The lower bound is further reduced by the compensation of the visual and sensor streams. A clear sky is favourable for stable rooftop power generation according to historical power and irradiance data. Partial Occlusion Requires Segmentation to Some Extent. During the missing-frame period, the edge-adaptive gate avoids adding old visual information to the sensor sequence. This behaviour is in line with the concept of conditional feature reliability [21], and the model should not give all input features the same weight under all circumstances.

The Computation Profile is also related. A segmentation-heavy model may have high accuracy but does not meet the edge latency requirement. The proposed Hiera-Seg design uses hierarchical regions and avoids dense full-resolution reasoning at all levels. Reduce redundant background processing and keep the forecast suitable for local gateways. Under resource constraints [22], only the essential modules and shadow evidence are retained, and non-essential background features are simplified.

### 4.3. Visual Occlusion Robustness and Review-action Interpretation

Table 1 shows the simulated orientation and occlusion scenarios used in the review-oriented evaluation. The table is small, but the mean power and visible occlusion rates of the records are different. In building microgrid assessment [23], the cloud-shade shift scenario has a lower mean power and higher occlusion uncertainty, while the rooftop array has a higher mean power and a lower occlusion rate. Therefore, the two forecasting models need to cover both the local shade event and the high-output stable period.

**Table 1. Simulated multi-orientation BIPV dataset setting.**

| Scenario | Records | Mean power kW | Occlusion rate |
|---|---|---|---|
| South facade | 3,900 | 38.4 | 0.21 |
| East facade | 3,500 | 24.8 | 0.33 |
| Rooftop array | 4,200 | 61.7 | 0.08 |
| Unseen building | 3,300 | 42.5 | 0.27 |
| Cloud-shade shift | 3,600 | 31.6 | 0.39 |

Figure 7 shows unseen-building robustness and calibration result. Figure 7(a) Unseen building tests whether the model adapts to a new facade or rooftop geometry without retraining all parameters. Figure 7(b) confidence calibration compares predicted confidence with actual error level and checks whether low-confidence windows correspond to difficult forecasts. Figure 7(c) shows missing-frame handling and tests whether the edge-adaptive gate relies more on sensors and historical power when visual input is unavailable. Figure 7(d) latency shift assesses whether stale image evidence still supports forecasting after edge processing delay [24].

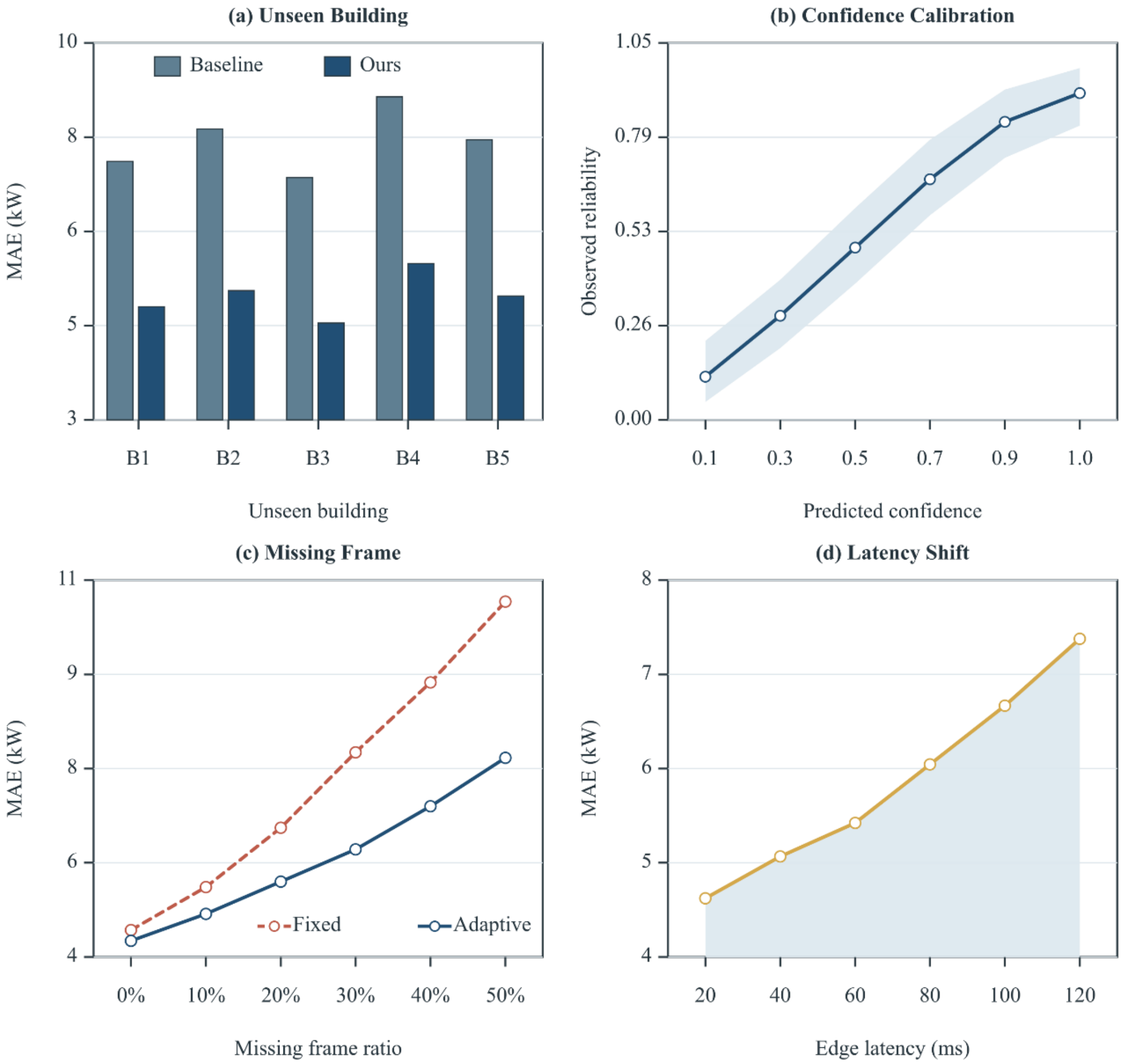

**Figure 7. Unseen-building robustness and calibration result. (a) Unseen Building. (b) Confidence Calibration. (c) Missing Frame. (d) Latency Shift.**

Robustness results show that the edge-adaptive gate improves the model with noisy data. When all the images and sensor streams are available, both the full model and the non-adaptive model perform well. If the visual frame is missing or delayed, the adaptive gate will not use the old segmentation evidence too much. Delayed visual evidence can mislead local controllers [25], so this behaviour is required for building energy-management systems.

Figure 8 shows the energy review card and dominant-evidence output. Figure 8(a) Review level converts forecast confidence and expected error into routine, watch, high-priority and manual-review categories. Figure 8(b) shows whether the forecast is mainly supported by irradiance, visual occlusion, orientation or recent power history. For reviewable energy forecasting [26], the two subfigures show both the predicted value and the evidence that caused the model to expect that value.

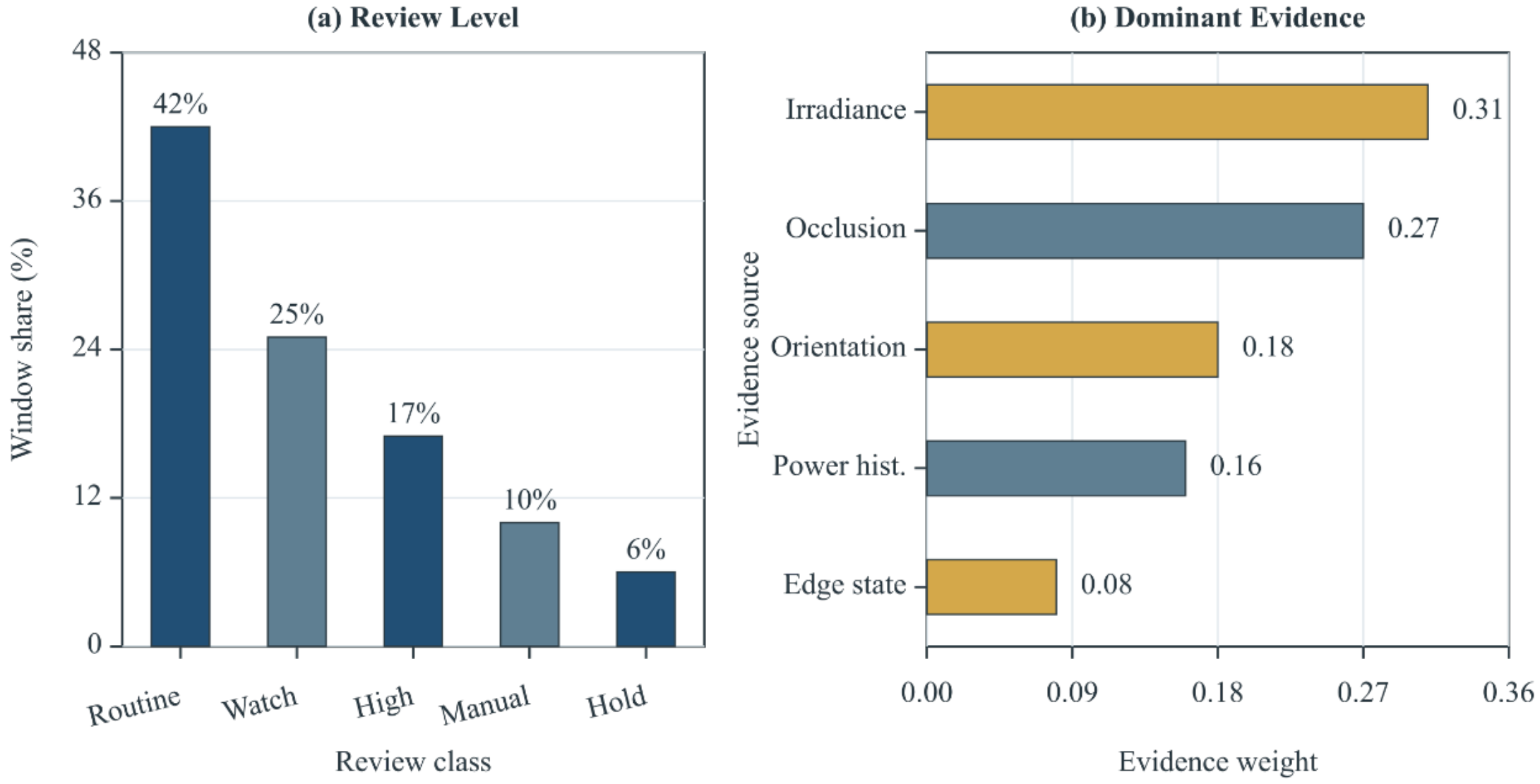

**Figure 8. Energy review card and dominant-evidence output. (a) Review Level. (b) Dominant Evidence.**

A review card is useful because BIPV forecasts often drive downstream energy decisions. A high-confidence rooftop forecast can support battery charging, while a low-confidence facade forecast under moving shadow may require conservative reserve. Dominant evidence also helps identify sensor problems: if a forecast is dominated by visual occlusion but contradicts irradiance and power history, the window can be flagged for manual inspection. This uncertainty-aware logic is consistent with scheduling practice [27].

The revised discussion acknowledges that the framework is not ready for unrestricted deployment. The simulated dataset does not cover all facade materials, inverter topologies, wiring mismatches and urban canyon geometries. Real BIPV images may contain reflections, dirt, bird droppings, maintenance scaffolds, seasonal vegetation and camera viewpoint drift. Before field deployment [28], segmentation-quality control, edge-device monitoring, routine calibration and comparison with real-meter data should be included in the operating workflow.

Despite these limitations, the results indicate that an edge-adaptive Hiera-Seg framework is suitable for multi-orientation BIPV forecasting. It connects visual occlusion with irradiance and power history, adapts feature fusion under edge-device uncertainty and generates reviewable output for building energy management. Because facade and rooftop modules behave differently [29], the method is most useful in locally shaded building areas where sensor-only models cannot represent spatial shading changes.

The application is also shown in the maintenance and operation plan. A repeated mismatch between segmentation evidence and power output can indicate dirty modules, sensor drift, inverter derating or an incorrectly mapped facade region. Therefore, the results of this review can be applied to prediction and determination of when a building energy model needs to be recalibrated. The above additional diagnostic information is needed to observe the long-term operating status of BIPV systems [30] in buildings with different facade colours and shade conditions over time.

The edge-oriented setting is more practical for smart buildings in the framework. Cloud-based inference is precise, but there will be communication and privacy delays. Local inference at the edge gateway can achieve a faster response for storage scheduling and demand-side control. If the model confidence is low, the controller will keep using the previous plan for now. This behaviour is in line with resilient energy management [31], and forecast uncertainty should be reflected in the downstream decisions.

Visual evidence is also relatively easy to learn from. A forecast may be correct but still hard to trust if the model cannot explain which facade or rooftop region caused it. The main-evidence card shows operators whether the forecast is affected by south facade occlusion, rooftop irradiance, edge-device delay or recent power history. As the model can be verified by visual and sensor data [32], it has relatively high operational acceptance compared with a hidden numerical predictor.

The model has not been trained on new buildings. The simulated results show good transfer; however, the real buildings may have irregular facade designs or module arrangements. The deployment strategy will start with a short calibration period, compare the predicted and actual power, tune edge-adaptive thresholds, and only then allow the forecast to control storage or load actions. Such staged deployment is common in building automation [33] because the operational risk is reduced by introducing data-driven control gradually.

In short, the results show that the proposed model has outperformed the baseline model. It is a visual-sensor explanation chain that shows module division, power prediction, and finally, review priority. This chain is suitable for BIPV applications because the generation source is physically visible on the building envelope. When review evidence is available [34], agreement or disagreement between the forecast and visible module state can also serve as an operational review.

The results also indicate that in the future, the benchmarks for BIPV forecasting will include visual occlusion, edge-device reliability and orientation-specific indicators. Average power error alone is not suitable for assessing a model used in building operation. A model may have a high total power but fail to represent facade groups that require local control. Review-oriented indicators [35] could be used to compare with actual energy-saving conditions.

## 5. Conclusion

This paper proposes an edge-adaptive Hiera-Seg model for building-integrated photovoltaic power forecasting. Align RGB facade images, optional thermal cues, irradiance measurements, module temperature, orientation labels, rooftop masks, shading labels, historical power and edge-device state in a method to create synchronous forecasting windows. Hiera-Seg extracts hierarchical module, shadow and background evidence, and the edge-adaptive gate adjusts feature fusion under missing frames, latency variation and sensor uncertainty. The model is designed to output the predicted power along with confidence and dominant evidence, and this output can be used in a review-oriented energy management workflow.

Based on the results of the simulation, the proposed framework is suitable for multi-directional BIPV forecasting under facade, rooftop, partial-shading and unseen-building scenarios. Based on the above experiments, visual occlusion, orientation state and irradiance dynamics are all necessary for stable short-term prediction. The dominant-evidence output is more suitable for the energy-management review, and operators can determine whether the predicted power is mainly due to irradiance, local shading, orientation or recent power history. It is suitable for storage scheduling or load shifting under a forecast, and a low-confidence value should not trigger an aggressive control operation.

Future work will use field BIPV data from buildings with different facade materials, inverter structures and urban shading environments. Deployment should include segmentation-quality assessment, edge-device monitoring and periodic recalibration based on measured power data. Additional studies should examine long-term facade soiling, seasonal vegetation change, camera viewpoint drift and operator override logs, because repeated overrides may indicate that visual evidence is correct but the scheduling policy needs adjustment.

## 6. Declarations

### 6.1. Author Contributions

Na Yu contributes to conceptualization, methodology, software, validation, analysis, investigation, data collection, draft preparation, manuscript editing, visualization, supervision, project administration, and funding acquisition. All authors have read and agreed with the manuscript before its submission and publication.

### 6.2. Funding and Acknowledgments

This research received no specific financial support from any funding agency.

### 6.3. Institutional Review Board Statement

Not applicable.

### 6.4. Informed Consent Statement

Not applicable.

### 6.5. Declaration of Competing Interest

Not applicable.

### 6.6. Declaration of Competing Interest

The authors declare that they have no known competing financial interests or personal relationships that could have appeared to influence the work reported in this paper.

### 6.7. Declaration of generative AI and AI-assisted technologies

During the preparation of this manuscript, the authors utilized Grammarly for linguistic polishing, logical organization and draft revision. The generative AI tool was not involved in research design, data collection, data analysis, or the equation of core conclusions. All outputs generated by AI were reviewed, revised and verified manually by the authors. The authors take full responsibility for all content of this manuscript.